\documentclass[11pt,leqno]{amsart}
\usepackage{amssymb}
\usepackage{graphics}
\usepackage{graphicx}
\usepackage{csquotes}
\usepackage{xspace}
\usepackage{booktabs,subcaption,colortbl}
\usepackage{tikz}
\usepackage{pgfplots}
\pgfplotsset{compat=1.18}

\newtheorem{thm}{Theorem}[section]
\newtheorem{crl}[thm]{Corollary}

\newtheorem{lmm}[thm]{Lemma}

\newtheorem{prp}[thm]{Proposition}

\theoremstyle{definition}
\newtheorem{dfn}[thm]{Definition}

\theoremstyle{remark}
\newtheorem{rem}{Remark}

\newcommand{\conv}{\operatorname{conv}}
\newcommand{\dist}{\operatorname{d}}
\newcommand{\rpd}{\operatorname{RPD}}
\newcommand{\NN}{\mathbb{N}}
\newcommand{\RR}{\mathbb{R}}
\newcommand{\Sph}{\mathbb{S}}
\newcommand{\cH}{\mathcal{H}}

\newcommand\SetOf[2]{\left\{#1\,\vphantom{#2}\right|\left.\vphantom{#1}\,#2\right\}}
\newcommand\smallSetOf[2]{\{#1 \,|\, #2\}}
\newcommand{\NCA}{\operatorname{NCA}}
\newcommand{\polymake}{\texttt{polymake}\xspace}
\newcommand{\oscar}{\texttt{OSCAR}\xspace}
\newcommand{\normaliz}{\texttt{Normaliz}\xspace}
\newcommand{\lmax}{\operatorname{\ell-max}}
\newcommand{\sd}{\operatorname{sd}}
\renewcommand\epsilon{\varepsilon}
\newcommand\abs[1]{\lvert #1 \rvert}

\title{Random Polytope Descriptors}

\author{Michael Joswig}
\address{Technische Universität Berlin \& MPI MiS Leipzig}
\email{joswig@math.tu-berlin.de}

\author{Marek Kaluba}
\address{Adam Mickiewicz University, Poznań, Poland}
\email{kalmar@mailbox.org}

\author{Lukas Ruff}
\address{AIgnostics}
\email{lukas@aignostics.com}


\begin{document}
%

\maketitle

\begin{abstract}      
  We introduce a class of random polytopes which simultaneously generalizes several known constructions.
  While being fairly general, these polytopes are also computationally exceptionally benign.
  We indicate how these properties can be exploited for classification and clustering tasks in data analysis.
  Crucially, our construction lets users smoothly trade off between a tighter description of the data and faster computation.
\end{abstract}

\section{Introduction}

Our goal is to present a class of random convex polytopes which are at the same time very general but also computationally benign; we call them \emph{random polytope descriptors}.
That is to say, these polytopes can approximate arbitrary polytopes somehow well enough; and at the same time obtaining a double description is feasible in practice, even for fairly large input.
In this way, the random polytope descriptors are useful for classification and clustering tasks in data analysis.
To fully understand the context and our contribution a few explanations are in order.

A polytope $P$ can be described as the convex hull $P=\conv(V)$ of a finite point set $V\subset\RR^d$ or as the bounded intersection $P=\bigcap \cH$ of a finite family of affine halfspaces $\cH$.
This basic fact has been called the \enquote{Main Theorem} of polytope theory \cite[Theorem 1.1]{Ziegler:Lectures+on+polytopes}.
A pair $(V,\cH)$ is a \emph{double description} of $P$.
The task of computing $\cH$ from $V$ is known as the \emph{convex hull problem}; the reverse task, i.e., computing $V$ from $\cH$, is the \emph{dual convex hull problem}.
Via cone polarity these two algorithmic problems are, in fact, equivalent.
A double description is necessary for computing combinatorial information about $P$, but it is generally useful for any number of geometric operations on polytopes.
Now it may happen that the coding lengths of $V$ and $\cH$ (say, if $V$ and $\cH$ are rational) differ considerably.
A first example is the $d$-dimensional unit cube $[0,1]^d$, which requires $2^d$ points but only $2d$ inequalities.
Taking these $2d$ inequalities as the input, the double description is exponentially large.
And the resulting computational challenge is a major obstacle in applying advanced methods from computational geometry to data science, where the input is often large.
The reader is referred to the monographs of Ziegler \cite{Ziegler:Lectures+on+polytopes}, Schrijver \cite{Schrijver:1986}, Boyd and Vandenberghe \cite{boyd2004} for background on polyhedral geometry and optimization; for empirical data on convex hull computation see \cite{ABS:1997,polymake:2017}.

Constructions of random polytopes and linear programs form an important and active line of research in polyhedral geometry and optimization; e.g., see B\'ar\'any \cite{Barany:1992,Barany-corrigendum}, Reitzner \cite{Reitzner:2004} and Newman \cite{Newman:2006.07000}.
A model case of a computationally benign class of random polytopes are convex hulls of random points on the unit sphere $\Sph^{d-1}$ or, dually, polytopes whose facet defining hyperplanes are tangent to $\Sph^{d-1}$; see Borgwardt \cite{Borgwardt:1987,Borgwardt:2007}.
However, without modification, these polytopes are too constrained to approximate many other polytopes too well.
At the core of Borgwardt's analysis is the rotational invariance of $\Sph^{d-1}$; we will follow that idea.

The motivation for the present work originates in data analysis.
More precisely, we are interested in settings in which the data is composed of different unobserved groups (clusters or classes), and where the joint distribution of observed variables is a mixture of the group-specific distributions.
This setting might be exemplified by assuming that the data come from a mixture of (latent) multivariate multinomial distributions, which is the natural assumption in many machine learning tasks such as clustering or classification.
We further assume that the super-level sets of the probability density function for each group-specific distribution in the mixture form a convex set, except maybe for regions of very low probability.
Relating ideas from convexity to machine learning and data analysis is standard; see, e.g., Vapnik~\cite{vapnik1995nature}, Sch\"olkopf and Smola \cite{scholkopf2002}; see also Cevikalp and Sa{\u{g}}lamlar \cite{cevikalp2021polyhedral} for one of several random constructions.

Our method is designed to work well on embeddings into feature space, where each of the classes is embedded into mutually disjoint convex sets.
Such an assumption is at the foundation of classical methods, e.g., support vector machines \cite{scholkopf2002}.
These methods, however, by simply partitioning the entire feature space among classes, do not take the finiteness of training data into account, i.e., most classes are usually given unbounded regions.
While such a feature space might seem desirable, as it affords generalization properties for models under a closed set assumption (assuming that only $K$ clusters or classes exist), it also increases the surface for adversarial and out-of-distribution attacks in open set situations \cite{scheirer2012}.
Here we attempt to remedy this limitation by explicitly partitioning the feature space into compact convex regions and one unbounded (non-convex) \enquote{out-of-distribution} set.

This article is structured as follows.
In Section~\ref{sec:polytopes} we recall dual bounding bodies and define our \emph{random polytope descriptors} as a variation.
An important side aspect is the discussion (in Section~\ref{sec:gauges}) of how to measure the distance from such a polytope in practice.
Our main theoretical contribution is Corollary~\ref{cor:barycenter} which provides a fast way of approximating the vertex barycenter of a dual bounding body (or a random polytope descriptor).
This observation ties in with practical feasibility of our computations.
Section~\ref{sec:learning} is dedicated to a machine learning experiment.
First we check the performance of the random polytope descriptors for classifying data from autoencoders.
Secondly, we analyze one class from one autoencoder experiment in greater detail.
The text closes with a brief outlook.

\subsection*{Acknowledgments}
MJ is indebted to RIMS for their hospitality during the \enquote{Conference on Discrete Optimization and Machine Learning (Kyoto, 2025)}.
We thank two anonymous referees for their comments and questions.
Claude Sonnet (versions 4.6 and 5) was used for two tasks: setting up the computational experiment in Section \ref{subsec:setup} and finally checking the overall consistency of this text.
This consistency check lead to adding Lemma~\ref{lem:direction-vertex} to give a more detailed proof of Theorem~\ref{thm:number_of_vertices}.
\normaliz \cite{bruns2017normaliz}, \oscar \cite{OSCAR-book}, and \polymake \cite{polymake:2017} were used for computations.
Funded by the Deutsche Forschungsgemeinschaft (DFG, German Research Foundation) under Germany's Excellence Strategy -- \enquote{The Berlin Mathematics Research Center MATH$^+$} (EXC-2046/1, EXC-2046/2, pro\-ject ID 390685689), \enquote{Symbolic Tools in Mathematics and their Application} (TRR 195, project ID 286237555), \enquote{Mathematical Research Data Initiative (MaRDI)} (project ID 460135501).

\section{Polytopes, Randomization, and Algorithms}
\label{sec:polytopes}
We want to explore some polyhedral geometry from a data analysis point of view.
Randomization will help us to maintain a tight control of the algorithmic complexity.
In this way, in Section~\ref{sec:learning} below, we will be able to study data sets of fairly high dimensions.

Consider a set $X \subset \RR^d$, typically finite and affinely spanning.
This is taken as input to an algorithmic pipeline which is tasked to return a convex body which somehow approximates $P=\conv(X)$, but subject to some complexity constraints.
Typically, the latter constraints do not allow to compute all facets of $P$, e.g., because the output can be prohibitively large.
Practical examples arise in real time computing where answers are required within a fixed time frame.

\subsection{Dual bounding bodies}
\label{sec:dual-bounding}
It is a standard idea to replace convex hulls by suitable dual bounding bodies, which form outer approximations.
Let $\Sph^{d-1}$ denote the $(d{-}1)$-dimensional sphere of unit vectors in $\RR^d$.
Throughout the following we assume that $Y\subset \Sph^{d-1}$ is a finite set.  

  The \emph{dual bounding body} of $X$ with respect to $Y$ is the polyhedron
  \begin{equation}\label{eq:dual-bounding}
    D_Y(X) \ = \ \SetOf{ v \in \RR^d }{ \langle v, y \rangle \leq \sup_{x\in X} {\langle x, y \rangle} \text{ for } y\in Y } \enspace.
  \end{equation}
By construction $D_Y(X)$ is convex, and it contains $X$; so it also contains
\begin{equation}\label{eq:conv}
  P \ := \ \conv(X) \ = \SetOf{\sum_{i=1}^{n} \lambda_i x_i}{x_i \in X,\, 0\leq\lambda_i,\, \sum\lambda_i=1} \enspace .
\end{equation}
If $X$ is finite, then the supremum in \eqref{eq:dual-bounding} is a maximum attained for some $x\in X$, which only depends on $y$.
In that case $D_Y(X)=P$ if and only if for each facet of $P$ the set $Y$ contains a normal vector of that facet.
Checking if $x\in\RR^d$ is contained in $D_Y(X)$ takes $O(\abs{Y}\cdot d)$ time.
In our analysis the sets $X$ and $Y$ will receive different roles; for a simple terminology we call the elements of $X$ \emph{points} and the elements of $Y$ \emph{directions}.

In the sequel we will assume that both, $X$ and $Y$, are finite, and that $Y$ is \emph{positively spanning}, i.e., $\conv(Y)$ contains the origin in its interior.
As $X$ is finite, the latter property is satisfied if and only if the polyhedron $D_Y(X)$ is bounded, i.e., it is a polytope.
Gauging the size of $Y$ gives us a handle on controlling the computational complexity.

\subsection{Random polytope descriptors}
\label{sec:rpd}
We have a data science scenario in mind where the input, $X$, is considered \enquote{noisy}, meaning that maybe some points in the input are not meant to be there in the first place.
Often the dual bounding bodies may be too large, as they contain the full input.
As a remedy, we propose a slightly more general class of convex polytopes.
Their algorithmic efficiency and robustness to the presence of outliers can be controlled by a pair of parameters.
Again we assume that $X\subset\RR^d$ is finite with $n=\abs{X}$.

\begin{figure}[htb]
  \centering
  \includegraphics[width=0.7\columnwidth]{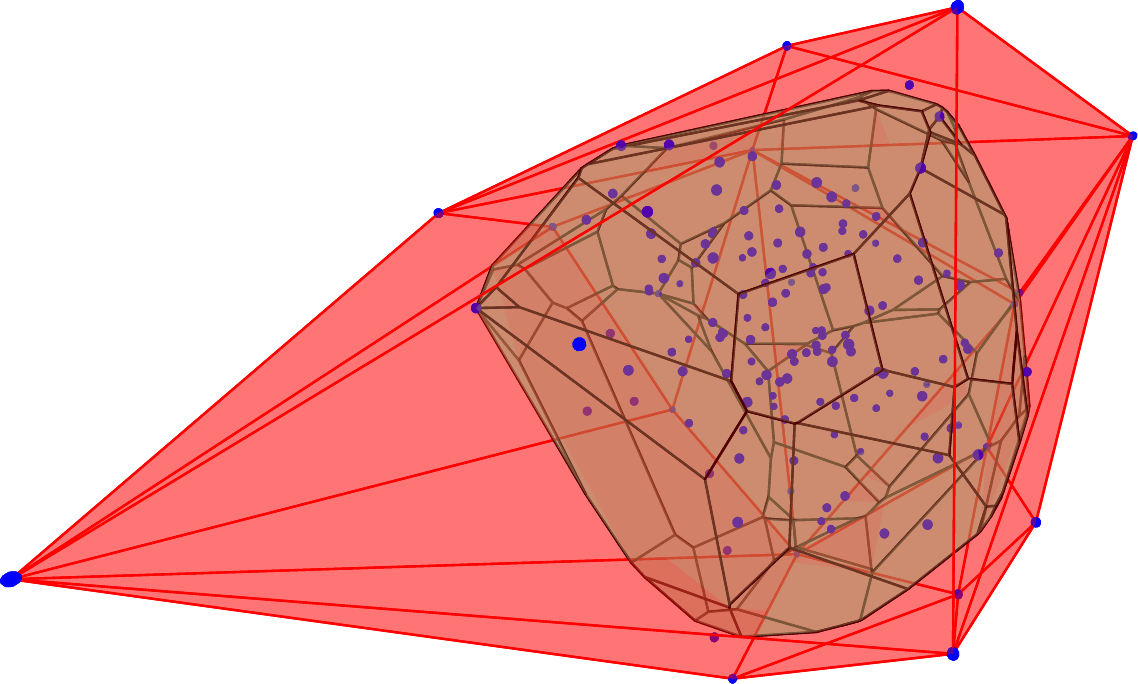}
  \caption{Convex hull $\conv(X)$ (red) and random polytope descriptor $\rpd_{80,3}(X)$ (green).
    Here $d=3$, and $X \sim N(0,I)$ is a random Gaussian sample of $150$ points with mean $0$ and identity covariance $I$.
    Despite $3\%$ outliers, the descriptor closely approximates the convex hull of the normal class (i.e., the inner cluster).}
  \label{fig:rpd}
\end{figure}

\begin{dfn}\label{def:rpd}
  Let  $Y\subset \Sph^{d-1}$ be a set of $m$ directions chosen uniformly at random, and let $\ell$ be a positive integer.
  The \emph{random polytope descriptor} is the polyhedron
  \begin{equation}\label{eq:rpd}
    \rpd_{m, \ell}(X) \ := \  \SetOf{ v\in \RR^d }{ \langle v, y \rangle \leq  \lmax_{x\in X} \langle x , y \rangle,\; y \in Y }\enspace ,
  \end{equation}
  where $\lmax$ denotes the $\ell$-th largest scalar product, and $\operatorname{1-max}=\max$.
\end{dfn}
Tacitly we will assume that $\rpd_{m, \ell}(X)$ is bounded.
This is the case if and only if $Y$ is positively spanning.
Since the $m$ directions in $Y$ are drawn i.i.d.\ and uniformly at random, this holds with high probability: by Wendel's theorem \cite{Wendel:1962}, the probability that $Y$ fails to be positively spanning is exactly $2^{-m+1}\sum_{k=0}^{d-1}\binom{m-1}{k}$, which tends to $0$ exponentially fast in $m$, for $d$ fixed.
The dual bounding bodies arise as special cases of random polytope descriptors; more precisely, $\rpd_{m, 1}(X)=D_Y(X)$.
The example in Figure~\ref{fig:rpd} captures the essence of our intuition behind the Definition~\ref{def:rpd}.

\begin{rem}
  In the context of unsupervised anomaly detection it is often beneficial to discard a fixed amount of data from $X$, i.e., label them as anomalous.
  The $\lmax$ function serves as a \enquote{threshold} to eliminate the most extreme points from the training set.
  This is similar in spirit to \emph{convex hull peeling}, where outliers are removed by iteratively stripping away the vertices of the convex hull; see Calka and Quilan \cite{CalkaQuilan:2023} for a recent probabilistic analysis.
  For (semi-)supervised learning, we note there may be natural candidate functions to replace $\lmax$ to define variants of $\rpd$, e.g., based on density estimation, weights, the value of a random variable, or simply the label.
\end{rem}

\begin{rem}
  If $Y$ happens to comprise all facet directions of $P=\conv(X)$, then we have $P=D_Y(X)=\rpd_{m, 1}(X)$.
  So, for a suitably chosen (non-generic) set of directions $Y$, every polytope arises as a random polytope descriptor; in particular, all combinatorial types occur.
  However, since $Y$ consists of $m$ i.i.d.\ continuously distributed directions, such a special alignment with the facet normals of $P$ happens with probability zero.
  Generically, i.e., with probability one, $\rpd_{m,\ell}(X)$ is instead a simple polytope, for any $m\ge d$; see Section~\ref{subsec:simple}.
\end{rem}

We briefly explain what it means to compute $\rpd_{m, \ell}(X)$, and which cost this incurs.
First we need $m$ random unit vectors, which can be found in $O(m)$ time; see \cite{TAOCP:2}, \S3.4.1.E.6.
There is no reason to be overly exact here, so it is adequate to assume that the coefficients of the random vectors are constantly bounded, i.e., floating point numbers with a fixed number of bits.
Consequently, all our complexity estimates are expressed in the unit cost model.
Throughout, we take the input space dimension $d$ as a constant.
Given $X$, we then evaluate $mn$ scalar products to obtain a description in terms of inequalities (so called \emph{H-representation}) \eqref{eq:rpd} of $\rpd_{m, \ell}(X)$, resulting in a total complexity of $O(mn)$.

In practice, it makes sense to compute $\rpd_{m,\ell}(X)$ symmetrically, i.e., by also finding the minimum, not only the maximum, because it essentially does not incur extra cost.
In that case the resulting polyhedron is determined by $2\cdot m$ inequalities.

\subsection{Scaling distance and central points}
\label{sec:gauges}
Any convex polytope $Q$, with the origin in its interior, can be used to define a \enquote{convex distance function}, which is symmetric if and only if $Q$ is centrally symmetric.
It is known to be occasionally useful to explicitly allow for $Q$ not to be centrally symmetric; see Martini and Swanepoel \cite{MartiniSwanepoel:2004} for a survey.
And this is true for our scenario, too.
However, here we want to go one step further by removing the special role of the origin.

Let $Q\subset\RR^d$ be a full-dimensional convex polytope with arbitrary interior point $c$.
The \emph{scaling distance} of $x\in\RR^d$ to $Q$ with respect to $c$ is defined as
\begin{equation}\label{eq:scaling-distance}
  \sd_c(x, Q) \ := \ \min\SetOf{ \alpha\geq 0 }{ x \in \alpha(Q - c) + c } \enspace .
\end{equation}
This is the smallest number $\alpha\geq 0$ such that $Q$ inflated by a factor of $\alpha$ from the \emph{central point} $c$ contains $x$.
The point $x$ is contained in $Q$ if and only if $\sd_c(x,Q)\leq 1$.

If $Q$ is a random polytope descriptor (as in Definition~\ref{def:rpd}), the scaling distance can serve as a dissimilarity (anomaly) score with respect to a given finite point set.
In this case, the points further away from $Q$ are assigned higher scores (see Section~\ref{sec:learning} below).
Given $x$, $c$, and an H-representation of $Q$ in terms of $m$ linear inequalities, we can compute $\sd_c(x, Q)$ in $O(m)$ time, for $d$ constant, by evaluating just scalar products.
In our data analysis scenario the choice of $c$ is crucial, because it defines the \enquote{ideal representative} of the input.

There are several natural candidates for the central point $c$ of a polytope $Q$, which differ with respect to computational complexity, when $Q$ is given in terms of inequalities.
For instance, the \emph{centroid} is the center of gravity of $Q$; this is hard even to approximate, see \cite{Rademacher:2007}.
The \emph{vertex barycenter} is the average of the vertices of $Q$; this is hard to compute exactly, see \cite{ElbassoniTiwary:2009};
further results in the latter article make it unlikely that there is an efficient approximation procedure either.
The \emph{Chebyshev central point} is the center of the largest \emph{sphere inscribed} to $Q$; this can be computed in $O(\sqrt{m})$ time by linear programming; see \cite{EavesFreund:1982} and \cite{Renegar:1988}.
Note that the latter worst case complexity assumes that the bit representation of each inequality in the H-representation of $Q$ is constantly bounded.
This is not an issue here in the sense that constantly bounded bit size is in line with our general setup and assumptions.
For the special case of $Q=\rpd_{m,\ell}(X)$, where $X$ is a normally distributed random sample, all three central points discussed are arbitrarily close to the origin, with high probability.
We say that a property, depending on $m\in\NN$, holds \enquote{with high probability} if it holds with probability converging to $1$ as $m$ approaches infinity.

A fourth candidate is the \emph{sample mean}, which is the average over the given point set $X$.
Notice that it may happen that the sample mean is not contained in $Q=\rpd_{m,\ell}(X)$ for $\ell>1$, i.e., if outliers have been discarded.
Yet, in the normally distributed setting, the sample mean is also arbitrarily close to the origin, with high probability. 
In Section \ref{sec:learning} we will use the sample mean due to its simplicity and observed better performance in higher dimensions.
However, Theorem~\ref{thm:number_of_vertices} below will provide a theoretical argument in favor of the vertex barycenter.

\subsection{Vertices and their barycenters}
\label{sec:vertices}

In this section we show that random polytope descriptors (as in Definition~\ref{def:rpd}) are exceptionally benign from a computational point of view.
To understand the context, we first recall basic facts from computational polyhedral geometry.
In \eqref{eq:rpd} the random polytope descriptor $\rpd_{m, \ell}(X)$ is given in terms of inequalities, and the parameter $m$ allows us to control the size of that description.
For many nontrivial computations it is necessary to describe a polytope in terms of a \emph{double description}, i.e., in terms of facets \emph{and} vertices.
Computing the facets of the convex hull of given points is known as the \emph{convex hull problem}; there are many algorithms and their implementations.
The dual problem, to pass from inequalities to vertices, is equivalent in view of cone polarity.
If the input is large, in practice, there are two fundamental obstacles.

The first obstacle is a combinatorial explosion.
Suppose $P=\conv(X)\subset\RR^d$ and $n=\abs{X}$.
Letting $f$ denote the number of facets of $P$, McMullen's upper bound theorem \cite{McMullen:1970} (see also \cite[\S8.4]{Ziegler:Lectures+on+polytopes}) says that
\begin{equation}\label{eq:upper-bound}
  f \in O\left(n^{\lfloor d/2\rfloor}\right) \,, \quad \text{if $d$ is considered constant.}
\end{equation}
That bound is actually tight as can be seen, for example, from cyclic polytopes.
In view of cone polarity, the same (tight) estimate holds for the number of vertices if the number of inequalities is given.
This means that the number of vertices and the number of facets of a polytope may differ by several orders of magnitude.
In practice, for given $X$, choosing a suitable parameter $m$ for the random polytope descriptor offers a tradeoff between a correct but expensive description of $P=\conv(X)$ and a loose but affordable one.

The second obstacle is an algorithmic one.
It is not known if there is a convex hull algorithm which runs in \emph{total polynomial time}.
That is, it is not known if the $f$ facets can be computed from the $n$ vertices in time which is bounded by a polynomial in $n$, $f$ and $d$.
For an overview concerning theoretical results and empirical data see \cite{ABS:1997} and \cite{polymake:2017}.
In the interest of a simplified complexity analysis, so far we often considered $d$ constant.
In that case there is a theoretically optimal algorithm \cite{Chazelle:1993}.
However, its worst case running time is dictated by the upper bound \eqref{eq:upper-bound}, which is polynomial in $d$, for $d$ constant, but still large.
Consequently, most convex hull computations are extremely hard or infeasible in practice for $d\geq 20$.
This means that for a typical data analysis application computing a full convex hull is usually impossible.
In practice, the \enquote{optimal} algorithm from \cite{Chazelle:1993} is usually much inferior to other algorithms \cite{ABS:1997,polymake:2017}, which are not polynomial time in the worst case, if $d$ is fixed.

The random polytope descriptors are so useful because they allow us to overcome both obstacles simultaneously, under the standard hypothesis that the input data is normally distributed with unknown mean and variance.
The following observations are not only relevant for data analysis.
In fact, they tie in with ongoing research on random polytopes theory \cite{Buchta1985,Reitzner:2010,Newman:2006.07000}.
We make use of Borgwardt's \emph{Rotation-Symmetry Model (RSM)} for random polytopes \cite{Borgwardt:1987}, which implies the following.
\begin{prp}\label{prp:slender}
  Let $n=\abs{X}$, and let the points of $X\subset\RR^d$ be drawn i.i.d.\ in some rotationally invariant way;
  the Gaussian distribution $N(0,I)$ or the uniform distribution from $\Sph^{d-1}$ form examples.
  Fix a positive integer~$\ell$.
  If $n=n(m)\to\infty$ sufficiently fast as $m\to\infty$, then, for $d$ constant, the number of vertices of $\rpd_{m,\ell}(X)$ is of order\/ $\Theta(m)$, with high probability.
\end{prp}
\begin{proof}
  By rotational invariance, for each fixed direction $y\in\Sph^{d-1}$ the scalar products $\langle x,y\rangle$, $x\in X$, have a common distribution $\mu$ not depending on~$y$.
  Hence $\lmax_{x\in X}\langle x,y\rangle$ is, for fixed $y$, the $\ell$-th order statistic from the top of $n$ i.i.d.\ draws from $\mu$, and classical order statistics theory \cite{DavidNagaraja:2003} yields a deterministic sequence $c_n\to\infty$ with
  \begin{equation}\label{eq:lmax}
    \lmax_{x\in X}\langle x,y\rangle = c_n\,(1+o_P(1)) \quad \text{ as $n\to\infty$, for $\ell$ fixed.}
  \end{equation}
  Here $o_P(\cdot)$ is the stochastic little-o; i.e., \eqref{eq:lmax} says that the ratio $\bigl(\lmax_{x\in X}\langle x,y\rangle\bigr)/c_n$ tends to one in probability.  
  A union bound over the (finitely many, and independent of $X$) directions $y_1,\dots,y_m\in Y$ shows that, provided $n$ grows fast enough with $m$, all $m$ offsets equal $c_n(1+o_P(1))$ simultaneously, with probability tending to~$1$.
  So, with high probability, $\rpd_{m,\ell}(X)$ is combinatorially indistinguishable from the classical constant-offset random polytope obtained by intersecting $m$ halfspaces tangent to a sphere of radius $c_n$ in the directions $Y$; the (routine but tedious) verification that this vanishingly small perturbation of the offsets does not change the vertex count we omit.
  By a result of Buchta, M\"uller, and Tichy \cite{Buchta1985}, see also \cite[Theorem~1]{Newman:2006.07000}, the number of vertices of that constant-offset polytope is $F(d)\,m\,(1+o(1))$ in expectation, for an explicit dimensional constant $F(d)>0$.
  Newman \cite[Theorem~12]{Newman:2006.07000} upgrades this to a high-probability statement via an Efron--Stein/Chebyshev argument.
  This proves the claim.
\end{proof}
Proposition~\ref{prp:slender} says that random polytope descriptors have very few vertices, with high probability, in sharp contrast to the worst case scenario described by the upper bound theorem~\eqref{eq:upper-bound}.
More precisely, they form a \emph{slender family} in the sense of \cite{Newman:2006.07000}.
Technically, the above result is not strictly necessary for the rest of the paper.
However, we believe that it helps the reader picture how and why random polytope descriptors \enquote{work}; see also Section~\ref{subsec:simple} below.

\begin{rem}
  The hypothesis $n=n(m)\to\infty$ in Proposition~\ref{prp:slender} is essential, not merely technical.
  If instead $X$ (hence $n$) is kept fixed while $\ell$ is fixed and $m\to\infty$, then $\lmax_{x\in X}\langle x,y\rangle$ is a fixed piecewise-linear function of $y\in\Sph^{d-1}$ with finitely many pieces, determined by $X$ and $\ell$ alone.
  Consequently $\rpd_{m,\ell}(X)$ converges, as $m\to\infty$, to the fixed polytope obtained by intersecting the halfspaces over \emph{all} $y\in\Sph^{d-1}$, and its number of vertices is then bounded by a constant depending on $n,\ell,d$ only, not on $m$.
  Growing $n$ together with $m$ avoids this saturation phenomenon.
\end{rem}

Next we want to discuss approximating the vertex barycenter of an $\rpd$.
This is meant to complement the discussion in Section~\ref{sec:gauges}.
We start out with a general lemma.
\begin{lmm}\label{lem:direction-vertex}
  Let $K\subset\RR^d$ be a convex body with $K\supseteq P=\conv(X)$, and suppose every point of $K$ is at distance at most $R$ from the origin.
  Then, for every $u\in\Sph^{d-1}$ and every $v\in\arg\max_{w\in K}\langle w,u\rangle$,
  \[
    \|v-u\| \ \leq \ \sqrt{R^2-1+2h_0}\enspace ,
  \]
  where $h_0:=\dist_H(P,\Sph^{d-1})$ is the Hausdorff distance.
\end{lmm}
\begin{proof}
  Since $\dist_H(P,\Sph^{d-1})=h_0$, there is $p\in P$ with $\|p-u\|\leq h_0$.
  As $\|u\|=1$,
  \[
    \langle p,u\rangle \ = \ \langle u,u\rangle - \langle u-p,u\rangle \ \geq \ 1-\|u-p\| \ \geq \ 1-h_0\enspace .
  \]
  Since $K\supseteq P$, the support value $h_K(u) = \langle v,u\rangle$ satisfies $h_K(u)\geq\langle p,u\rangle\geq 1-h_0$.
  Using $\|v\|\le R$,
  \[
    \|v-u\|^2 \ = \ \|v\|^2 - 2\langle v,u\rangle + \|u\|^2 \ \leq \ R^2-2(1-h_0)+1 \ = \ R^2-1+2h_0 \enspace . \qedhere
  \]
\end{proof}
This sharpens the reduction sketched in the appendix of \cite{Newman:2006.07000} --- which was itself posed in response to an earlier version of the present work --- to a precise, deterministic bound.

To keep the exposition concise we now focus on the case where $X$ is uniformly sampled from $\Sph^{d-1}$, which is the easiest RSM type; this leads to the random polytopes $P(d,n)$.
We say that $P \sim P(d, n)$ if $P = \conv(X)$, where $X\subset\Sph^{d-1}$ comprises $n$ points chosen uniformly at random.
Since $X$ consists of $n$ i.i.d.\ continuously distributed points on the sphere, $P$ is simplicial, with $n$ vertices, with probability one; Borgwardt \cite{Borgwardt:2007} further shows that, with high probability, its $\Theta(n)$ many facets can be computed in $O(n^2)$ time, and the vertex barycenter lies arbitrarily close to the origin, for $d$ constant.
Note that here $D_X(X)$ agrees with the \emph{polar dual} $P^\vee:=\smallSetOf{v\in\RR^d}{\langle v,x\rangle \leq 1,\; x\in X}$.
\begin{thm}
  \label{thm:number_of_vertices}
  Let $P=\conv(X)\sim P(d, n)$ as above, with $h_0:=\dist_H(P,\Sph^{d-1})$.
  Let $Y\subset \Sph^{d-1}$ be a set of $m$ directions chosen uniformly at random.
  Fix $\epsilon>\sqrt{2h_0}$ and $0<p<1$.
  Then, for $m$ sufficiently large, the \emph{mean} of $s$ randomly chosen vertices of $D_Y(X)$ is at distance $\leq\epsilon$ from the origin with probability at least $1-p$ if
  \[
    s \ > \ \frac{2}{\bigl(\epsilon-\sqrt{2h_0}\bigr)^2}\left(1+\frac2d\log\frac2p\right)\enspace .
  \]
\end{thm}

\begin{proof}
  The condition $\dist_H(P, \Sph^{d-1}) \leq h$, where $\dist_H$ is the Hausdorff distance, may be read as: the height of the largest spherical cap cut from $\Sph^{d-1}$ by some facet of $P$ is at most~$h$.
  So we have
  \[
    \dist_H(D_Y (X), \Sph^{d-1}) \ \leq \ h(d,m,p')\enspace ,
  \]
  with probability $1-p'$, where $h(d,m,p')$ is the distance resulting from the worst case scenario of all points from $Y$ missing the largest spherical cap induced by $P$, and $p':=p/2$.
  Formally,
  \[
    h(d,m,p') := \inf\SetOf{h}{\binom{m}{d} \left(1-\NCA_d(h)\right)^{m-d}\leq \frac{p'}{2}} \enspace ,
  \]
  where $\NCA_d(h)$ is the area of a spherical cap of height $h$ normalized so that $\NCA_d(1) = 1$ regardless of dimension.
  Write $h:=h(d,m,p')$; all facets, hence all vertices, of $D_Y(X)$ then lie at distance at most $1+h$ from the origin, with probability at least $1-p'$.

  Fix $u_1,\dots,u_s\in\Sph^{d-1}$ i.i.d.\ uniformly at random, and let $v_i\in\arg\max_{v\in D_Y(X)}\langle v,u_i\rangle$ be the corresponding randomly sampled vertex, for $i=1,\dots,s$; the same sampling procedure is used below in the proof of Corollary~\ref{cor:barycenter}.
  Conditionally on the event above, Lemma~\ref{lem:direction-vertex}, with $K=D_Y(X)$ and $R=1+h$, gives, deterministically,
  \[
    \|v_i-u_i\| \ \leq \ \delta \ := \ \sqrt{(1+h)^2-1+2h_0} \qquad (i=1,\dots,s)\enspace .
  \]
  The triangle inequality yields
  \[
    \Bigl\|\tfrac1s\textstyle\sum_{i=1}^s v_i\Bigr\| \ \leq \ \Bigl\|\tfrac1s\textstyle\sum_{i=1}^s u_i\Bigr\| + \delta\enspace .
  \]
  By \cite[Lemma~21]{Newman:2006.07000}, applied to the i.i.d.\ uniform points $u_1,\dots,u_s$ on the unit sphere ($R=1$),
  \[
    \Bigl\|\tfrac1s\textstyle\sum_{i=1}^s u_i\Bigr\| \ \leq \ \epsilon-\delta
  \]
  holds with probability at least $1-p'$, provided $s > \frac{2}{(\epsilon-\delta)^2}\bigl(1+\frac2d\log\frac1{p'}\bigr)$.
  As $m\to\infty$, $h\to0$ and hence $\delta\to\sqrt{2h_0}$; taking $m$ large enough that $\delta<\epsilon$ and combining the two events we obtain the probability at least $1-p'-p'=1-p$.
  This proves the claim.
\end{proof}
Now let $X$ be uniformly distributed on the translated sphere $v+\Sph^{d-1}$.
Then the vertex barycenter of $D_Y(X)$ is close to $v$, with high probability.
The following result says that $v$ can be recovered from $D_Y(X)$ quickly, and the running time does not depend on the cardinality of $X$.
\begin{crl}\label{cor:barycenter}
  Let\/ $D_Y(X)$ be given in terms of $m$ linear inequalities.
  Then its vertex barycenter can be approximated, with high probability, in $O(\sqrt{m})$ time, if $d$ is fixed.
\end{crl}
\begin{proof}
  By Theorem~\ref{thm:number_of_vertices} it suffices to know a constant number of random vertices of $D_Y(X)$ to approximate the vertex barycenter, for given $\epsilon$ and $p$.
  Solving a linear program, with the $m$ inequalities describing $D_Y(X)$ as constraints and a random direction as the objective function, provides one random vertex.
  Each such linear program can be solved in $O(\sqrt{m})$ time \cite{Renegar:1988}.
\end{proof}

\begin{rem}
Standard results on the RSM \cite{Borgwardt:1987} suggest that a version of Corollary~\ref{cor:barycenter} also holds for $X$ normally distributed with $v$ as its unknown mean, and for $\rpd_{m,\ell}(X)$ instead of $D_Y(X)$, albeit at a slightly higher cost.
The somewhat technical details are beyond our current scope.
\end{rem}

Corollary~\ref{cor:barycenter} implies that the scaling distance of any point to a random polytope descriptor with respect to the vertex barycenter can be approximated extremely fast, e.g., for normally distributed input with unknown mean and variance.

\section{Experiments}
\label{sec:learning}

Here we use random polytope descriptors to explore the geometry of feature spaces learned by deep autoencoders.
In the following experiments, we compare and use $\rpd$ as a quality measure to examine and validate the convexity assumption in the latent space of established deep autoencoder models.
Our code and data are available at Zenodo, \texttt{doi:10.5281/zenodo.22913313}.

\subsection{A classification scenario}
\label{subsec:classification}
We consider $K$ distinct finite sets $X_1,X_2,\dots,X_K\subset\RR^d$, where $X_k$ is viewed as the set of $n_k$ samples of class~$k$, seen in some feature space $\RR^d$.
Our goal is to describe how to classify a new point $x\in\RR^d$ with respect to $X_1,X_2,\dots,X_K$ as training data.

Associated with these $K$ classes are the convex polytopes $P_k = \conv(X_k)$.
If all pairwise intersections $P_k\cap P_j$ are disjoint for $k \neq j$, then these polytopes provide perfect descriptors, provided that we have an oracle to test whether a new sample $x\in\RR^d$ is contained in one of the sets $P_1,P_2,\dots,P_K$ or their complement.
Checking if $x$ lies in $P_k$ can be decided via a linear program in dimension $n_k$, with $O(d+n_k)$ constraints, and hence in $O(\sqrt{n_k})$ time.
Replacing $P_k$ by the random polytope descriptor $\rpd_{m,\ell}(X_k)$ changes that complexity to $O(m)$.
For practical considerations, it should be noted that our very coarse complexity analysis hides fairly large constants for linear programming.
Moreover, the strong sublinear bound $O(\sqrt{n_k})$ for linear programming is lost if data needs to be copied prior to the computation.

To avoid cumbersome notation we now write $P_k=\rpd_{m,\ell}(X_k)$, i.e., we replace $\conv(X_k)$ by a suitable RPD.
%
Let $X = \bigcup_k X_k$ be the training data partitioned into classes.
For the \emph{estimated distance} of a point $x \in \RR^d$ to the combined support of all classes we take the minimum
\begin{equation}\label{eq:distance-to-support}
  \delta(x) \ := \ \min_k \bigl\{ \sd_{c_k}(x, P_k) \bigr\}
\end{equation}
of the scaling distances, where $P_k = \rpd_{m,\ell}(X_k)$ and $c_k$ is a central point.
Note that $\delta(x)$ is a random variable as the construction of polytopes $P_k$ is not deterministic.
If the estimated distance to the combined support is larger than one, then evaluating \eqref{eq:distance-to-support} assigns a class which is closest to $x$, namely the index $k$ at which the minimum is attained.
For generic $x$ this is unique.
It follows from our complexity analysis above that estimating the distance to the combined support takes $O(m)$ time, for $d$ and $K$ fixed.
This assumes that the central points $c_1,\dots,c_K$ are known.
For the sample means that cost is linear in the size of~$X$; see Section~\ref{sec:vertices}.

All $\rpd$ computations in this section use the Chebyshev center as the central point; see Section~\ref{sec:outlook} for a comment on the vertex barycenter.

\subsection{Experimental setup}
\label{subsec:setup}
In our experiments we use the known MNIST \cite{mnist} and Fashion-MNIST (FMNIST) \cite{fmnist} 
image datasets, each containing $K=10$ classes.
The autoencoder training has been organized with the help of Claude Sonnet 4.6 in the Claude Code CLI.
We trained a vanilla autoencoder (AE) and a variational autoencoder (VAE).
Both autoencoders operate on flattened $28\times 28$ images ($784$ input units) and share the same encoder backbone and decoder.

AE is a three-layer fully-connected network $784 \xrightarrow{\mathrm{ReLU}} 512 \xrightarrow{\mathrm{ReLU}} 256 \to d$, where the final layer is linear and $d$ denotes the latent dimension.
The decoder mirrors this structure: $d \xrightarrow{\mathrm{ReLU}} 256 \xrightarrow{\mathrm{ReLU}} 512 \xrightarrow{\sigma} 784$, with a sigmoid output.
The network is trained to minimize the binary cross-entropy reconstruction loss, averaged over the batch.
At inference time the encoder output is used directly as the embedding~\cite{ruff2018}.                                               
                                                                                                                                        
The VAE backbone ($784 \to 512 \to 256$, ReLU) feeds two parallel linear heads that produce the posterior mean $\mu \in \RR^d$ and log-variance $\log\sigma^2 \in \RR^d$.                                                                                         
During training, a latent code is sampled via the reparameterization trick, $z = \mu + \sigma \odot \varepsilon$, $\varepsilon \sim \mathcal{N}(0, I)$, and passed to the same decoder as the AE.                                                                                             
The training objective is the evidence lower bound (ELBO):                                                                            
\[                                                                                                                                    
  \mathcal{L} = \mathbb{E}[\log p(x \mid z)]                                                                                          
  - D_{\mathrm{KL}}\bigl(\mathcal{N}(\mu,\sigma^2 I) \,\|\, \mathcal{N}(0,I)\bigr),                                       
\]                                                                                                                                    
where the first term is the binary cross-entropy reconstruction loss and the KL divergence has the closed form $-\tfrac{1}{2}\sum_j(1 + \log\sigma_j^2 - \mu_j^2 - \sigma_j^2)$~\cite{kingma2014}.
At inference time the posterior mean $\mu$ is used as the embedding (no sampling).                                                    
Both models are trained with the Adam optimizer~\cite{kingmaAdam2015} (learning rate $10^{-3}$, batch size $256$) for $30$ epochs.
None of the networks are given any explicit convexity prior (apart from the Gaussian prior for VAE).

\begin{table}[tb]
\caption{AUC scores of $\rpd_{640, 1}$ and $k$-means for AE and VAE models trained on MNIST (classes $0$--$9$ in (a)) and FMNIST (classes $F0$--$F9$ in (b)) with embedding dimensionality $d=20$.
We set $\ell=1$ in this experiment, as the $\rpd$s are trained in a supervised setting, and thus no pollution of the training data is expected.
Justification for choosing $m=640$ is given in sensitivity analysis in Figure~\ref{fig:kmeans_comparison_auc}.}
\label{tab:kmeans_comparison_auc}
\begin{center}
\begin{footnotesize}
\begin{subtable}{.5\linewidth}
	\caption{MNIST}
	\centering
	\renewcommand{\arraystretch}{1.33}
	\rowcolors{2}{gray!0}{gray!05}
	\begin{tabular}{c c c c c c}
	\toprule
	& \multicolumn{2}{c}{AE} && \multicolumn{2}{c}{VAE}\\\cmidrule{2-3}\cmidrule{5-6}
	\rowcolor{gray!0} Class & $\rpd$ & $k$-means && $\rpd$ & $k$-means\\\midrule
	$0$ & \textbf{97.9} & 97.3 && \textbf{95.6} & 89.7\\
	$1$ & 98.5 & \textbf{98.8} && 96.7 & \textbf{97.1}\\
	$2$ & 92.6 & \textbf{95.7} && 86.1 & \textbf{86.5}\\
	$3$ & 93.1 & \textbf{95.1} && \textbf{91.9} & 87.6\\
	$4$ & 93.9 & \textbf{95.7} && \textbf{90.5} & 84.0\\
	$5$ & 91.5 & \textbf{91.6} && \textbf{89.9} & 81.1\\
	$6$ & 96.1 & \textbf{97.1} && \textbf{92.8} & 88.7\\
	$7$ & 94.8 & \textbf{95.7} && \textbf{91.5} & 90.5\\
	$8$ & 93.2 & \textbf{93.5} && \textbf{93.4} & 86.9\\
	$9$ & \textbf{94.9} & 91.7 && \textbf{92.3} & 83.6\\
	\bottomrule
	\normalsize
\end{tabular}
\end{subtable}%
\begin{subtable}{.5\linewidth}
	\caption{FMNIST}
	\centering
	\renewcommand{\arraystretch}{1.33}
	\rowcolors{2}{gray!0}{gray!05}
	\begin{tabular}{c c c c c c}
	\toprule
	& \multicolumn{2}{c}{AE} && \multicolumn{2}{c}{VAE}\\\cmidrule{2-3}\cmidrule{5-6}
	\rowcolor{gray!0} Class & $\rpd$ & $k$-means && $\rpd$ & $k$-means\\\midrule
	$F0$ & 86.0 & \textbf{94.4} && \textbf{83.6} & 83.6\\
	$F1$ & 88.5 & \textbf{97.7} && 80.5 & \textbf{89.3}\\
	$F2$ & 85.1 & \textbf{90.0} && \textbf{84.0} & 80.0\\
	$F3$ & 89.8 & \textbf{93.5} && \textbf{81.6} & 81.0\\
	$F4$ & 86.2 & \textbf{90.1} && 75.9 & \textbf{79.5}\\
	$F5$ & 87.4 & \textbf{90.0} && \textbf{82.5} & 78.1\\
	$F6$ & 83.4 & \textbf{83.5} && \textbf{79.9} & 72.2\\
	$F7$ & 92.3 & \textbf{97.6} && 84.7 & \textbf{93.6}\\
	$F8$ & 81.8 & \textbf{91.3} && 75.6 & \textbf{81.3}\\
	$F9$ & 90.5 & \textbf{97.8} && 85.2 & \textbf{87.2}\\
	\bottomrule
	\normalsize
\end{tabular}
\end{subtable}
\end{footnotesize}
\end{center}
\end{table}

\subsection{Comparing RPD to $k$-means}
\label{subsec:auc}
We first compare $\rpd$ to a common $k$-means classifier which finds the mean of each class and then assigns a score to each sample based on the distance to the class mean.
The resulting classifier, in geometric terms, partitions the latent space into Voronoi regions of the class means; see \cite{Polyhedral+and+Algebraic+Methods} for a discussion of Voronoi diagrams and algorithms.
In contrast, the $\rpd$ classifier is based on the scaling distance to the class polytopes with respective sample means.
We train the classifiers on the AE and VAE embeddings of the MNIST and FMNIST training sets, respectively, and evaluate the separation of each class on the full test sets using the common Area Under the ROC curve (AUROC or AUC) metric.

While $\rpd$ and $k$-means are similar in spirit, they have significant differences.
Due to the non-isotropic nature of the scaling distance, $\rpd$ should achieve a tighter convex description than $k$-means, especially when clusters in feature space are not rotation invariant.
This is true for VAE networks, for example, if the latent space dimension is too low as clusters become packed.
Therefore, we expect $\rpd$ to have fewer false positives at the same false negative rates, or equivalently, $\rpd$ to achieve better AUCs in this experiment; see Table~\ref{tab:kmeans_comparison_auc} and Figure~\ref{fig:kmeans_comparison_auc}.
VAE $\rpd$ (with $d=20$) wins over $k$-means for 8 out of 10 classes, while $k$-means dominates for FMNIST AE across the board; VAE $\rpd$ for FMNIST has comparable results between the two classifiers.

Another key advantage of $\rpd$ is the canonical rejection threshold given by the scaling distance, which can be understood as assigning another \enquote{unknown} class to low density areas in the latent space.
A threshold to replicate such behavior for a $k$-means classifier would need to be found a posteriori (e.g., must be learned), but this increases the complexity and may result in a bias.

By no means do we claim that $\rpd$ yields optimal classifiers out of the box; instead it tends to pick up convexity in the input data, which may be useful.
For instance, Table~\ref{tab:kmeans_comparison_auc} shows that $\rpd$ works rather nicely with VAE embeddings.

\begin{figure}[p] \centering
  \newcommand\scalefactor{.55}

  \input{rpc-auc-diagrams-30epochs.tex}

  \caption{AUC scores of $\rpd_{m,1}$ (with $m\in\{80,160,320,640\}$) vs $k$-means average over all classes (black dashed curve).  MNIST data set (10 classes);  AE left, VAE right}
  \label{fig:kmeans_comparison_auc}
\end{figure}

\subsection{A simple RPD case study}
\label{subsec:simple}
RPD is attractive as a replacement for a direct convex hull computation, because it is possible to \enquote{choose} the amount of compute that the user is willing to spend in order to obtain an approximation with certifiable properties.
This section comprises one additional experiment to illustrate what this means.

First, we note that constructing the H-representation \eqref{eq:rpd} of the $\rpd$ is very fast in practice.
For instance, from the autoencoder data (MNIST/AE) we build a set $X$ of 6000 random points in $\RR^{20}$.
Computing $\rpd_{640,2}(X)$, with respect to the Chebyshev central point, takes between 0.13 and 0.14 seconds, depending on the class.
Most of the time is spent on computing the inequalities (about 85--90\% of total time); computing the Chebyshev center is fast and consistent at about 0.02 seconds.
Evaluating the scaling distance \eqref{eq:scaling-distance} is extremely fast, roughly at 40 microseconds per call.
The value of $\ell=2$ does not really matter here and below.

\begin{table}[th]
  \centering
  \begin{tabular}{lrrrrr}
    \toprule
    & \multicolumn{5}{c}{Class} \\
    \cmidrule(lr){2-6}
    & 0 & 1 & 2 & 3 & 4 \\
    \midrule
    Wallclock (s) & 341     & 390     & 380     & 395     & 419     \\
    usr (s)       & 863.03  & 989.39  & 960.39  & 986.66  & 1051.90 \\
    sys (s)       & 1580.69 & 1791.77 & 1789.87 & 1871.54 & 1911.73 \\
    CPU (s)       & 2443.72 & 2781.16 & 2750.26 & 2858.20 & 2963.63 \\
    \midrule
    & 5 & 6 & 7 & 8 & 9 \\
    \midrule
    Wallclock (s) & 403     & 369     & 372     & 376     & 366     \\
    usr (s)       & 1043.25 & 942.26  & 950.75  & 971.57  & 938.63  \\
    sys (s)       & 1877.29 & 1710.19 & 1715.35 & 1733.68 & 1691.73 \\
    CPU (s)       & 2920.54 & 2652.45 & 2666.10 & 2705.25 & 2630.36 \\
    \bottomrule
  \end{tabular}
  \caption{Timing breakdown by class for $\rpd_{640,1}$ exact dual convex hull computation (MNIST data set with VAE, embedding dimension $d=6$).  Mean wallclock time $381.1$, standard deviation $20.7$. Convex hull code: \normaliz v3.10.3 \cite{bruns2017normaliz}, with six parallel threads, run through \polymake v4.15 \cite{polymake:2017}.  The high proportion of sys time comes from interprocess communication.}
  \label{tab:dual-hull-timings}
\end{table}

Secondly, we want to understand the tradeoff between the speed of computation and the approximation accuracy.
For this experiment again we pick $n=6000$ points from the MNIST/AE data set.
However, to be computationally viable, we pick a much smaller embedding dimension $d=6$.
We convert the floating point coordinates to exact rational numbers, and compute the vertices of $\rpd_{640,1}$ for each of the ten classes; the timings are given in Table~\ref{tab:dual-hull-timings}.
One important observation is that it is possible to get an exact double description within reasonable time.
The reason why this is possible is Proposition~\ref{prp:slender}, which tells us that an RPD has very few vertices.
For applications where verified output is essential, this property is crucial but difficult to obtain by other means.
Further, the timings vary very little with the individual classes.
That is, the running time primarily depends on the parameter $m=640$, which can be chosen arbitrarily.

Of course, the latter freedom of choice comes at a price, which we explain next.
The particular choice of $d=6$ and $n=6000$ allows us to compute the convex hull $P=\conv(X)$ directly.
With \normaliz v3.10.3 (six parallel threads) via \polymake v4.15 that computation requires 730 wallclock seconds (for class 0).
Driving the parameters considerably larger would render this experiment infeasible; cf.\ \cite{polymake:2017}.
The polytope $P$ is simplicial (as expected), and its $f$-vector reads
\[
  643 \quad 13144 \quad 68797 \quad 143886 \quad 131385 \quad 43795 \enspace ;
\]
in particular, the ratio of the number of inequalities for the RPD versus the true number of facets of the convex hull equals $\tfrac{2\cdot 640}{43795}\approx 2.9\%$.
Recall that we compute RPD symmetrically; hence the factor two in the numerator.
On the other hand, $Q=\rpd_{640,1}(X)$ (with exact rational coordinates) is simple with $f$-vector
\[
  123161 \quad 369483 \quad 401382 \quad 186959 \quad 33121 \quad 1222 \enspace .
\]
By construction $P$ is contained in $Q$, which, however, is much larger than $P$.
So the message is that $\rpd$ works rather well as a classifier, at least for the data analyzed in Table~\ref{tab:kmeans_comparison_auc}, with very few inequalities.
Almost all of the $2\cdot 640=1280$ inequalities turn out to be facet defining.

In this example, both polytopes, $P$ and $Q$, are computationally benign in the sense that the number of facets of $P$ (as a function of $d$ and $n$) and the number of vertices of $Q$ (as a function of $d$ and $m$) are several orders of magnitude smaller than predicted by the upper bound theorem \eqref{eq:upper-bound}.


All timings in this section have been taken on an ASUS desktop (Pro WS X570-ACE motherboard), AMD Ryzen 9 5900X (12-core, 24 threads via SMT), running at up to ~4.95 GHz; 128 GB main memory.
Operating System: openSUSE Leap 15.6, Linux kernel 6.18.6, x86-64; \oscar \cite{OSCAR-book}: version 1.7; Julia: version 1.12.5; Perl: version 5.26.1.


\section{Conclusion and Outlook}
\label{sec:outlook}

The very little variance of the running times for the various classes in Section~\ref{subsec:auc} suggests that RPD may be suitable for real-time computations.
Yet the main selling point of the random polytope descriptors is the possible tradeoff between tight descriptions of data sets and computational speed, at the user's discretion.
Tight descriptions are particularly useful for defending against out-of-distribution attacks in classification scenarios.




In Section~\ref{sec:polytopes} we advertised the vertex barycenter.
Corollary~\ref{cor:barycenter} tells us that (in fixed dimension) the complexity for approximating the vertex barycenter is asymptotically the same as computing the Chebyshev center.
However, the constants work in favor of the Chebyshev center: in the setting of Section \ref{subsec:auc} $(d=20, m=640, \ell=2)$ computing the average of 148 random vertices of $\rpd$ takes between 0.97 and 1.00 seconds; this is more expensive than the Chebyshev center by a factor of eight.
To put this into perspective, it should be noted that for most polytopes in the same regime approximating the vertex barycenter is infeasible in practice.
So, once again, $\rpd$ offers a flexible framework to construct classifiers for scenarios where some amount of convexity among the input is expected.
It is easy to imagine scenarios where the vertex barycenter yields a better classifier than the Chebyshev center.
For all of the above, relevant tests are much desirable, but beyond our current scope.


\bibliographystyle{amsplain}
\bibliography{AEPolytopes}

\end{document}